\documentclass[conference]{IEEEtran}
\IEEEoverridecommandlockouts
\usepackage{cite}
\usepackage{amsmath,amssymb,amsfonts}
\usepackage{algorithmic}
\usepackage{graphicx}
\usepackage{textcomp}
\usepackage{array}
\usepackage{booktabs}
\usepackage{arydshln}
\usepackage{adjustbox}
\usepackage{xcolor}
\usepackage{url}
\def\BibTeX{{\rm B\kern-.05em{\sc i\kern-.025em b}\kern-.08em
    T\kern-.1667em\lower.7ex\hbox{E}\kern-.125emX}}
\begin{document}
\title{Language-Conditioned Visual Grounding with CLIP Multilingual}

\author{
	\IEEEauthorblockN{J. de Curtò\IEEEauthorrefmark{1}\IEEEauthorrefmark{2}\IEEEauthorrefmark{5}, Mauro Liz\IEEEauthorrefmark{2}\IEEEauthorrefmark{3},  I. de Zarzà\IEEEauthorrefmark{4}\IEEEauthorrefmark{5}}
	\IEEEauthorblockA{\IEEEauthorrefmark{1}\textit{Department of Computer Applications in Science \& Engineering}, 
	\textit{BARCELONA Supercomputing Center}, 
	Barcelona, Spain}
\IEEEauthorblockA{\IEEEauthorrefmark{2}\textit{Escuela Técnica Superior de Ingeniería (ICAI)}, 
\textit{Universidad Pontificia Comillas}, 
Madrid, Spain \\
Email: jdecurto@icai.comillas.edu, mauroliz@bu.edu}
\IEEEauthorblockA{\IEEEauthorrefmark{3}\textit{Department of Electrical and Computer Engineering}, 
	\textit{Boston University}, 
	Boston, MA 02215, USA}
\IEEEauthorblockA{\IEEEauthorrefmark{4}\textit{Human centered AI, Data \& Software}, 
	\textit{LUXEMBOURG Institute of Science and Technology}, 
	Esch-sur-Alzette, Luxembourg \\
	Email: irene.zarza@list.lu}
\IEEEauthorblockA{\IEEEauthorrefmark{5}\textit{Estudis d'Informàtica, Multimèdia i Telecomunicació}, 
	\textit{Universitat Oberta de Catalunya}, 
	Barcelona, Spain}	
}

\maketitle

\begin{abstract}
Multilingual vision--language models exhibit systematic performance gaps
across languages, but the mechanism remains ambiguous: cross-language
divergence could arise from the visual encoder, the text branch, or their
interaction. We resolve this ambiguity through a dense multilingual CLIP
probe in which the visual encoder is held \emph{identical} across thirteen
typologically diverse languages and only the XLM-RoBERTa text branch
varies. We evaluate two CLIP architectures spanning a $7\times$
visual-encoder scale gap (XLM-R base + ViT-B/32, $\sim$87\,M visual
parameters; XLM-R large + ViT-H/14, $\sim$632\,M) on 11 concepts and
210~images, and quantify cross-language agreement via cluster-mask IoU,
top-percentile IoU, and SPEARMAN rank correlation against an English
reference ($n=2{,}310$ paired observations per language). Three findings
emerge. First, low-resource languages (Arabic, Basque, Luxembourgish)
incur a structural penalty at both backbone scales (Wilcoxon HR\,$>$\,LR
$p<10^{-300}$; cluster-mask IoU gap $+0.114$ at base, $+0.143$ at large),
isolating the deficit to the text branch. Second, scaling the encoder
$7\times$ \emph{widens} the gap for structural failure cases (Basque
$\Delta=-0.056$, Luxembourgish $\Delta=-0.076$) while improving Arabic
($\Delta=+0.033$), separating corpus-coverage from
tokeniser-fertility failures. Third, peak similarity is preserved across
languages (mean ratio $0.94$ at large scale) while cluster-mask IoU
drops sharply, identifying \emph{spatial misalignment}, not signal
collapse, as the dominant failure mode. At $3.4$--$3.9$\,Wh per 1{,}000
queries, dense-CLIP grounding is competitive with high-throughput
inference budgets, positioning it as a practical substrate for
energy-aware multilingual deployment.
\end{abstract}



\section{Introduction}
\label{sn:intro}

Multilingual vision--language models (VLMs) are increasingly deployed in
settings where linguistic diversity is a first-class operational
requirement: cross-border autonomous systems, multinational
search-and-rescue platforms, content-moderation pipelines that span
dozens of languages, and assistive technologies for non-English-speaking
users. The dominant architectures couple a visual encoder, typically a
Vision Transformer pretrained contrastively on web-scale image--text
pairs, with a multilingual text branch, either through a frozen
contrastive backbone such as multilingual
CLIP~\cite{radford2021clip} or through an autoregressive language model
fine-tuned on visual instruction
data~\cite{liu2024llava16,chen2024internvl2,wang2024qwen2vl,abdin2024phi3}.
Both paradigms inherit a long-documented inequity from their text-only
predecessors: performance degrades systematically on languages
underrepresented in pretraining
corpora~\cite{hu2020xtreme,conneau2020unsupervised,ahuja2023mega,lai2023chatgpt},
with low-resource languages and non-Latin scripts paying the steepest
cost.

The mechanism behind this multilingual penalty is, however, ambiguous.
In an end-to-end VLM, a degradation observed on Arabic or Basque inputs
may originate in (a) the visual encoder responding differently to
non-English instruction tokens, (b) the multilingual text branch itself
producing weaker cross-lingual representations, (c) the autoregressive
decoder generating less coherent output in low-resource languages, or
(d) any interaction among these. Tokeniser fertility
disparity~\cite{rust2021good,petrov2024language} provides one
mechanistic candidate, but it acts on the text side and does not, on
its own, predict the spatial structure of perceptual failures. The
existing literature establishes that a multilingual penalty exists
without localising it to a specific stage of the model.

A second, orthogonal difficulty concerns how the penalty is
\emph{measured}. Existing multilingual VLM evaluations typically score
generated outputs against predominantly-English reference vocabularies
or
embeddings~\cite{ahuja2023mega,lai2023chatgpt,reimers2019sentence},
which structurally penalise non-English responses regardless of
whether the underlying visual reasoning is correct \cite{DECURTO2026104878}. The penalty
reported in the literature \cite{dronesdeCurto2026} thus aggregates two distinct effects, a
genuine cross-language perceptual disparity and a measurement bias
toward English, whose relative contribution cannot be recovered from
the published numbers.

This paper addresses both difficulties through a controlled probe: a
dense multilingual CLIP grounding pipeline~\cite{radford2021clip} in
which the visual encoder is held \emph{identical} across thirteen
typologically diverse languages, and cross-language agreement is
quantified between dense similarity maps using metrics that are
symmetric in their two arguments. Holding the visual encoder constant
isolates the penalty to the text branch; symmetric scoring removes the
English-anchored measurement artefact. The penalty we measure is
therefore attributable to the multilingual encoder itself, not to a
specific visual backbone or to a measurement framework biased toward
English.

We evaluate two CLIP architectures spanning a $7\times$ visual-encoder
scale gap (XLM-R~base + ViT-B/32, $\sim$87\,M visual parameters; XLM-R
large + ViT-H/14, $\sim$632\,M) on 11 BDD100K-relevant
concepts~\cite{yu2020bdd100k} across 210 images and 13 languages,
yielding $n = 2{,}310$ paired observations per non-English language.
Three findings emerge that, together, characterise the structure of
the multilingual penalty.

\begin{itemize}
    \item \textbf{The penalty persists when the visual encoder is held
          constant.} The high-resource versus low-resource cluster-mask
          IoU gap is $+0.114$ at the smaller backbone and $+0.143$ at
          the larger one, with Wilcoxon HR\,$>$\,LR significance below
          $10^{-300}$ at both backbone scales. This isolates the deficit to the
          text branch and rules out an interaction-only explanation.

    \item \textbf{Scaling the encoder $7\times$ widens, rather than
          closes, the gap for structural failure cases.} Basque
          ($\Delta\!=\!-0.056$) and Luxembourgish
          ($\Delta\!=\!-0.076$) lose IoU when the encoder is scaled,
          while Arabic ($\Delta\!=\!+0.033$) and Chinese
          ($\Delta\!=\!+0.039$) recover. This separates corpus-coverage
          failures (Basque, Luxembourgish) from
          tokeniser-fertility failures (Arabic, Chinese) on operational
          grounds.

    \item \textbf{The failure mode is spatial misalignment, not signal
          collapse.} Mean peak-similarity ratio relative to English is
          $0.94$ at the larger scale, peak signal is preserved across
          languages, while cluster-mask IoU drops sharply, indicating
          that the encoder activates on the wrong region rather than
          producing weak activation. The implication for downstream
          systems is that confidence-based filtering on peak similarity
          is insufficient to detect language-induced grounding errors.
\end{itemize}

A subsidiary energy result motivates dense-CLIP grounding as a
deployment substrate in its own right: at $3.4$--$3.9$\,Wh per 1{,}000
queries, the probe operates more than an order of magnitude below the
inference budget reported for autoregressive multilingual
VLMs~\cite{aienergyscore2025,luccioni2024power}, while delivering
spatially grounded predictions amenable to downstream reasoning. We
return to this trade-off in Section~\ref{sn:discussion}.

The remainder of the paper is organised as follows.
Section~\ref{sn:related} reviews multilingual evaluation of foundation
models, dense-feature CLIP grounding, and the energy-measurement
methodology. Section~\ref{sn:method} describes the probe pipeline,
metrics, and statistical protocol. Section~\ref{sn:results} reports the empirical findings across both
backbones, organised around three structural questions
(persistence, scale, mechanism).
Section~\ref{sn:discussion} interprets the results, discusses
implications for multilingual VLM design, and acknowledges limitations.
Section~\ref{sn:conclusion} concludes.


\section{Related Work}
\label{sn:related}

We organise the literature across three threads that converge in this
study: multilingual evaluation of foundation models
(Section~\ref{sn:rw_multilingual}), dense feature extraction from
contrastive vision--language backbones
(Section~\ref{sn:rw_dense_clip}), and energy efficiency of
foundation-model inference (Section~\ref{sn:rw_energy}). A short
positioning paragraph closes the section.

\subsection{Multilingual Evaluation of Foundation Models}
\label{sn:rw_multilingual}

The evaluation of large language models across typologically diverse
languages has a substantial but predominantly text-centric history. The
XTREME benchmark~\cite{hu2020xtreme} established a multilingual
protocol covering 40 languages and nine tasks, exposing systematic
cross-lingual gaps even in state-of-the-art models. XGLUE
\cite{liang2020xglue} extended this to generation tasks, while
XLM-R~\cite{conneau2020unsupervised} demonstrated that unsupervised
cross-lingual pretraining at scale could narrow but not eliminate the
gap between English and other languages. The MEGA evaluation~\cite{ahuja2023mega} and the ChatGPT multilingual study of
Lai et al.~\cite{lai2023chatgpt} confirmed that
performance disparities persist in instruction-tuned generative models
and correlate with training-corpus size per language; comparative
reasoning evaluations across foundation
models~\cite{deCurto2025_3} extend this to multi-step inference tasks
and find similar cross-model variability.

A structural source of cross-lingual inequity sits at the
\emph{tokeniser}. Rust et al.~\cite{rust2021good} showed that
multilingual BERT's subword vocabulary allocates tokens unevenly
across languages, causing morphologically rich languages to require
more tokens to express equivalent semantic content. Petrov et
al.~\cite{petrov2024language} systematised this observation as
tokeniser \emph{fertility disparity}: low-resource and non-Latin-script
languages consistently receive fewer vocabulary entries and thus higher
token-per-character ratios. Wendler et
al.~\cite{wendler2024llamas} complement these findings on the
representation side, showing that predominantly-English-pretrained
multilingual transformers perform internal reasoning in an
English-aligned latent space regardless of input language. Together,
these results identify the text branch as the natural locus of
multilingual bias, but they do so through text-only experiments. The
extent to which the same mechanisms govern multilingual \emph{visual}
grounding remains open.

For multimodal models specifically, multilingual evaluation is
substantially less developed than for text-only LLMs. Most VLM
benchmark papers report results in English only, with occasional
Chinese or bilingual evaluation, and the multilingual visual
question-answering literature has tended to operate via translation
pipelines rather than native multilingual inference. When multilingual
VLM evaluations are conducted, scoring metrics typically compare
generated outputs against English reference vocabularies or
embeddings~\cite{ahuja2023mega,lai2023chatgpt,reimers2019sentence},
which conflates genuine perceptual disparity with measurement bias
toward English, an artefact our probe is designed to avoid by scoring
in a symmetric, language-agnostic similarity-map space.

\subsection{Dense Feature Extraction from CLIP}
\label{sn:rw_dense_clip}

CLIP~\cite{radford2021clip} aligned image and text representations in a
shared embedding space through contrastive pretraining on web-scale
image--text pairs. The original formulation produces a single
image-level embedding suitable for classification and retrieval but not
for spatially grounded prediction. MaskCLIP~\cite{zhou2022maskclip}
showed that dense per-patch features can be extracted from a frozen
CLIP backbone by repurposing the final attention layer, enabling
zero-shot dense prediction without supervised retraining. This insight
has since been adopted in robot navigation, where systems such as
OneMap~\cite{busch2025onemap} use dense CLIP features as the substrate
for open-vocabulary mapping and language-conditioned goal localisation.
The OpenCLIP project~\cite{ilharco2021openclip} released a family of
multilingual checkpoints trained on
LAION-5B~\cite{schuhmann2022laion5b}, including the XLM-RoBERTa-text
variants we evaluate here, in which the visual encoder is shared
across all languages and only the text branch is language-conditioned.
This architectural property, a fixed visual stage with a swappable
multilingual text head, is what makes the present probe possible: the
visual representation is held identical across languages by
construction, not by approximation.

\subsection{Energy Efficiency of Foundation-Model Inference}
\label{sn:rw_energy}

Awareness of the environmental cost of deep learning has grown rapidly
since Strubell et al.~\cite{strubell2019energy} quantified the carbon
footprint of large transformer training. Schwartz et
al.~\cite{schwartz2020green} responded with a call for
\emph{Green AI}, treating computational efficiency as a first-class
publication criterion. Subsequent work refined these estimates by
accounting for hardware generation and grid carbon
intensity~\cite{patterson2021carbon,henderson2020energy,dodge2022measuring},
and the focus shifted from training to inference: Luccioni et
al.~\cite{luccioni2024power} provided the first systematic comparison
of inference energy across NLP tasks and architectures, finding that
generative models are substantially more expensive than discriminative
ones, and that energy per query varies by up to two orders of magnitude
across tasks. The AI Energy Score~\cite{aienergyscore2025} addresses
the resulting need for a standardised, hardware-agnostic
inference-efficiency metric, reporting energy in Wh per 1{,}000 queries
through continuous NVML-based GPU power sampling. We adopt the same
protocol here, enabling direct comparison between contrastive grounding
and autoregressive multilingual VLMs on a common footing. See Table~\ref{t:positioning} for a relative positioning of the contribution to closely related work.


\begin{table}[t]
\caption{Positioning of the present study relative to closely related
  work along five axes.
  \checkmark~= addressed; --~= not addressed.
  \emph{Multimodal}: jointly processes image and text inputs.
  \emph{Multilingual}: evaluates more than one language.
  \emph{Dense grounding}: produces spatially-resolved similarity maps
  rather than scalar scores or single-vector embeddings.
  \emph{Cross-lang.\ probe}: holds one model component constant
  while varying language, isolating the multilingual penalty
  to a specific architectural stage.
  \emph{Energy}: reports inference energy via standardised
  hardware-level measurement.}
\label{t:positioning}
\centering
\small
\adjustbox{max width=\linewidth}{%
\begin{tabular}{lccccc}
\toprule
\textbf{Study} & \textbf{Multimodal} & \textbf{Multilingual} &
  \textbf{Dense grounding} & \textbf{Cross-lang.\ probe} &
  \textbf{Energy} \\
\midrule
Hu et al.\ (2020)~\cite{hu2020xtreme}             & --        & \checkmark & --         & --         & --         \\
Conneau et al.\ (2020)~\cite{conneau2020unsupervised} & --        & \checkmark & --         & --         & --         \\
Ahuja et al.\ (2023)~\cite{ahuja2023mega}          & \checkmark & \checkmark & --         & --         & --         \\
Lai et al.\ (2023)~\cite{lai2023chatgpt}           & --        & \checkmark & --         & --         & --         \\
Petrov et al.\ (2024)~\cite{petrov2024language}    & --        & \checkmark & --         & \checkmark & --         \\
Wendler et al.\ (2024)~\cite{wendler2024llamas}    & --        & \checkmark & --         & \checkmark & --         \\
Zhou et al.\ (2022)~\cite{zhou2022maskclip}        & \checkmark & --        & \checkmark & --         & --         \\
Busch et al.\ (2025)~\cite{busch2025onemap}        & \checkmark & --        & \checkmark & --         & --         \\
Luccioni et al.\ (2024)~\cite{luccioni2024power}   & --        & --        & --         & --         & \checkmark \\
AI Energy Score (2025)~\cite{aienergyscore2025}    & \checkmark & --        & --         & --         & \checkmark \\
\midrule
\textbf{This work} & \checkmark & \checkmark & \checkmark & \checkmark & \checkmark \\
\bottomrule
\end{tabular}
}
\end{table}

\subsection*{Positioning}

Existing multilingual VLM benchmarks measure cross-language performance
gaps end-to-end, conflating contributions from the visual encoder, the
text branch, the autoregressive decoder, and the scoring metric.
Existing dense-CLIP work uses the architecture for grounding and
navigation but does not probe its multilingual structure. Existing
multilingual evaluations of CLIP focus on classification or retrieval,
not dense grounding. Our contribution sits at the intersection: a
controlled multilingual probe of dense CLIP grounding in which the
visual encoder is held identical across thirteen languages and only the
text branch varies, allowing the multilingual penalty to be localised
to a specific stage of the model and decomposed into a signal-strength
component and a spatial-alignment component.


\section{Method}
\label{sn:method}

The probe pipeline is designed to isolate the multilingual penalty
to the text branch of a vision--language model by holding the visual
encoder identical across all evaluated languages. Cross-language
agreement is then measured between dense similarity maps in a symmetric,
language-agnostic way, with no language privileged at scoring time. We
describe the formal probe construction, the experimental setup, the
metrics, and the statistical protocol in turn.

\subsection{Probe Construction}
\label{sn:probe}

Let $\mathcal{O} = \{O_1, \ldots, O_N\}$ denote a fixed set of $N$
images, $\mathcal{C}$ a set of concept terms (e.g., \emph{car},
\emph{pedestrian}), and $\mathcal{L}$ a set of natural languages with
$\ell_0 \in \mathcal{L}$ a designated reference (English). Let
$T_\ell : \mathcal{C} \to \mathbb{R}^d$ denote the language-$\ell$
text encoder mapping a concept to a dense embedding, and let $V$
denote a shared visual encoder mapping each image $O$ to a dense
feature map $V(O) \in \mathbb{R}^{H \times W \times d}$. For each
triplet $(O, c, \ell)$ the dense similarity map is
\begin{equation}
  S^{(\ell)}_{c}(O)[u,z]
  \;=\;
  \frac{V(O)[u,z] \cdot T_\ell(c)}
       {\|V(O)[u,z]\|\,\|T_\ell(c)\|}
  \;\in\;[-1,1]^{H\times W},
  \label{e:sim}
\end{equation}
defined pointwise over spatial locations $(u,z)$.

The probe holds $V$ \emph{identical} across all $\ell\in\mathcal{L}$:
the only quantity that varies between languages is the text encoder
$T_\ell$, which on multilingual CLIP architectures shares the same
output projection across languages but differs in its language-specific
parameter activations. Cross-language agreement on $(O, c)$ is the
agreement between $S^{(\ell)}_{c}(O)$ and $S^{(\ell_0)}_{c}(O)$ in the
spatial domain. Because both maps live in the same dense embedding
space and are produced by the same visual encoder applied to the same
image, the comparison is symmetric: no language is privileged at scoring
time, and the English-anchored keyword artefact common to autoregressive
multilingual VLM evaluation does not arise.

\subsection{Models, Dataset, Languages, Concepts}
\label{sn:setup}

\paragraph{Models.}
We evaluate two multilingual CLIP backbones from the
OpenCLIP project~\cite{ilharco2021openclip}, both pretrained on
LAION-5B~\cite{schuhmann2022laion5b} with an XLM-RoBERTa text
encoder~\cite{conneau2020unsupervised} (Table~\ref{t:models}). The
two architectures span a $7\times$ visual-encoder scale gap
($\sim$87\,M vs.\ $\sim$632\,M visual parameters), enabling the
scale-effect analysis of Section~\ref{sn:results}.

\begin{table}[t]
\caption{Multilingual CLIP backbones evaluated.}
\label{t:models}
\centering
\small
\adjustbox{max width=\linewidth}{%
\begin{tabular}{lll}
\toprule
\textbf{Short name} & \textbf{OpenCLIP id.} & \textbf{Visual params} \\
\midrule
XLM-R base + ViT-B/32 &
  \texttt{xlm-roberta-base-ViT-B-32}  & $\sim$87\,M \\
XLM-R large + ViT-H/14 &
  \texttt{xlm-roberta-large-ViT-H-14} & $\sim$632\,M \\
\bottomrule
\end{tabular}
}
\end{table}

\paragraph{Dataset.}
We sample $N = 210$ images from the BDD100K
corpus~\cite{yu2020bdd100k} via a deterministic frozen subset
(seed $= 42$) drawn jointly from train, validation, and test splits to
preserve distributional diversity. BDD100K provides scenes with
substantial variation in weather, lighting, road type, and object
density, which we use as a stress test for cross-language grounding
robustness rather than for any UAV-specific application.

\paragraph{Languages and concepts.}
We evaluate $|\mathcal{L}| = 13$ languages spanning six typological
families: Arabic (Semitic), Basque (isolate), Catalan, French, Italian,
Portuguese, Spanish (Romance), German, English, Luxembourgish
(Germanic), Russian (Slavic), and Chinese in both Simplified and
Traditional scripts (Sinitic). Three are explicitly tagged
low-resource: Arabic, Basque, Luxembourgish.
Concepts $\mathcal{C}$ comprise eleven BDD100K-relevant nouns:
\emph{car, truck, bus, person, pedestrian, traffic\_light,
traffic\_sign, bicycle, motorcycle, road, building}. The total count
of paired observations per non-English language is therefore
$N_\text{img} \cdot |\mathcal{C}| = 210 \cdot 11 = 2{,}310$.

\subsection{Dense Feature Extraction}
\label{sn:dense}

CLIP's visual encoder produces a single image-level embedding via
attention pooling at the final transformer layer. To recover dense
per-patch features suitable for spatial grounding, we follow the
MaskCLIP recipe of Zhou et al.~\cite{zhou2022maskclip}: the final
attention block's value projection is reapplied to the full sequence of
patch tokens (rather than only to the [CLS] token), yielding a dense
feature tensor of shape $H \times W \times d$ that respects the
training-time alignment with the text encoder. No fine-tuning of either
encoder is performed. The resulting dense maps are bilinearly
upsampled to $224 \times 224$ for the spatial-IoU comparison.

\subsection{Cross-Language Agreement Metrics}
\label{sn:metrics}

For a triplet $(O, c, \ell)$ with $\ell \neq \ell_0$, agreement is
computed against the English reference map $S^{(\ell_0)}_{c}(O)$ via
four complementary metrics.

\paragraph{Cluster-mask IoU (primary)}
We extract spatially coherent high-similarity regions from each
similarity map by region-growing on local maxima with a relative
threshold of $0.8$ and a minimum cluster size of $3$ patches, yielding
a binary mask $M^{(\ell)}_c(O) \subseteq [H]\times[W]$. The
cluster-mask IoU is
\begin{equation}
  \mathrm{IoU}_\text{cluster}^{(\ell)}(O, c)
  \;=\;
  \frac{|M^{(\ell)}_c(O) \cap M^{(\ell_0)}_c(O)|}
       {|M^{(\ell)}_c(O) \cup M^{(\ell_0)}_c(O)|}.
  \label{e:iou}
\end{equation}
This metric captures whether the two languages agree on the spatial
support of the concept, independently of absolute similarity scale.

\paragraph{Top-percentile IoU}
For a percentile $p \in \{90, 95, 99\}$, we threshold each map at
its $p$-th percentile and compute the IoU of the resulting binary
masks. This is a scale-invariant complement to cluster-mask IoU that
does not depend on the region-growing parameters.

\paragraph{SPEARMAN rank correlation}
$\rho^{(\ell)}(O, c)$ is computed pixel-wise between
$S^{(\ell)}_{c}(O)$ and $S^{(\ell_0)}_{c}(O)$, capturing whether the
two maps agree on the \emph{relative ordering} of all spatial
locations. This is a continuous, threshold-free agreement metric.

\paragraph{Peak-similarity ratio}
The peak-ratio statistic $\eta^{(\ell)}(O, c) = \max S^{(\ell)}_{c}(O)
/ \max S^{(\ell_0)}_{c}(O)$ measures whether the language-$\ell$ map
attains a comparable maximum activation to English. This metric
discriminates two failure modes: \emph{signal collapse}
($\eta\!\ll\!1$, low IoU) versus \emph{spatial misalignment}
($\eta\!\approx\!1$, low IoU). The two failures have different
operational implications for downstream confidence-based filtering.

\subsection{Statistical Protocol}
\label{sn:stats}

For each metric and each backbone we apply three complementary tests
on the $n = 2{,}310$ paired (image, concept) observations.

\paragraph{FRIEDMAN test across languages}
A FRIEDMAN non-parametric repeated-measures
test~\cite{friedman1937use} is applied with non-English languages as
treatments and (image, concept) pairs as blocks, testing whether
agreement differs systematically across languages.

\paragraph{Paired Wilcoxon HR\,$>$\,LR}
For each (image, concept) we compute the mean metric value across
high-resource (HR) and low-resource (LR) languages separately, then
test the one-sided alternative HR\,$>$\,LR via a paired WILCOXON
signed-rank test. This is the headline test for the multilingual
penalty.

\paragraph{Per-language Mann--Whitney}
For each low-resource language we test, via a one-sided
MANN--WHITNEY $U$ test, whether its metric distribution is dominated by
the pooled distribution over high-resource languages.

\subsection{Energy Measurement}
\label{sn:energy}

GPU energy is sampled at $10$\,Hz throughout the inference run via the
NVIDIA Management Library, following the AI Energy Score
methodology~\cite{aienergyscore2025}. All measurements are taken on an NVIDIA H200 (141\,GB HBM3e) on the Boston University HPC
cluster. Total energy is the time-integral
of instantaneous power over the measurement window; the reported
$E_{1\text{K}}$ value (Wh per 1{,}000 queries) normalises by the total
number of (image, concept, language) inferences. The protocol is
identical to that used by Luccioni et al.~\cite{luccioni2024power} and
permits direct comparison across architectures and against
autoregressive multilingual VLMs measured under the same protocol.


\section{Results}
\label{sn:results}

We organise the empirical findings around three structural questions
about the multilingual penalty: whether it is observable when the
visual encoder is held identical across languages
(Section~\ref{sn:o1}), whether scaling the encoder closes the gap
(Section~\ref{sn:o2}), and which failure
mechanism---signal collapse or spatial misalignment---dominates
(Section~\ref{sn:o3}). Headline statistics are summarised in
Table~\ref{t:summary}.

\begin{table}[t]
\caption{Headline per-backbone summary. HR = high-resource (9 non-English
  languages), LR = low-resource (Arabic, Basque, Luxembourgish).
  $n = 2{,}310$ paired (image, concept) observations per
  non-English language. $E_{1\text{K}}$ in Wh per 1{,}000 queries.}
\label{t:summary}
\centering
\small
\setlength{\tabcolsep}{4pt}
\begin{tabular}{lcc}
\toprule
\textbf{Statistic} & \textbf{Base} & \textbf{Large} \\
\midrule
Visual params (M)            & 87           & 632          \\
HR mean cluster-mask IoU     & 0.6123       & 0.6076       \\
LR mean cluster-mask IoU     & 0.4978       & 0.4647       \\
HR\,$-$\,LR IoU gap          & $+0.1145$    & $+0.1428$    \\
Wilcoxon HR\,$>$\,LR $p$     & $<10^{-300}$ & $<10^{-300}$ \\
Friedman $\chi^{2}$ (12 langs.) & 8065.8    & 9478.4       \\
Mean Spearman, HR            & 0.876        & 0.890        \\
Mean Spearman, LR            & 0.727        & 0.694        \\
Mean peak-ratio, HR          & 0.982        & 0.946        \\
Mean peak-ratio, LR          & 0.978        & 0.909        \\
$E_{1\text{K}}$ (Wh / 1K q.) & 3.87         & 3.36         \\
\bottomrule
\end{tabular}
\end{table}

\subsection{The Penalty Persists Under a Shared Visual Encoder}
\label{sn:o1}

\begin{figure}[t]
  \centering
  \includegraphics[width=\linewidth]{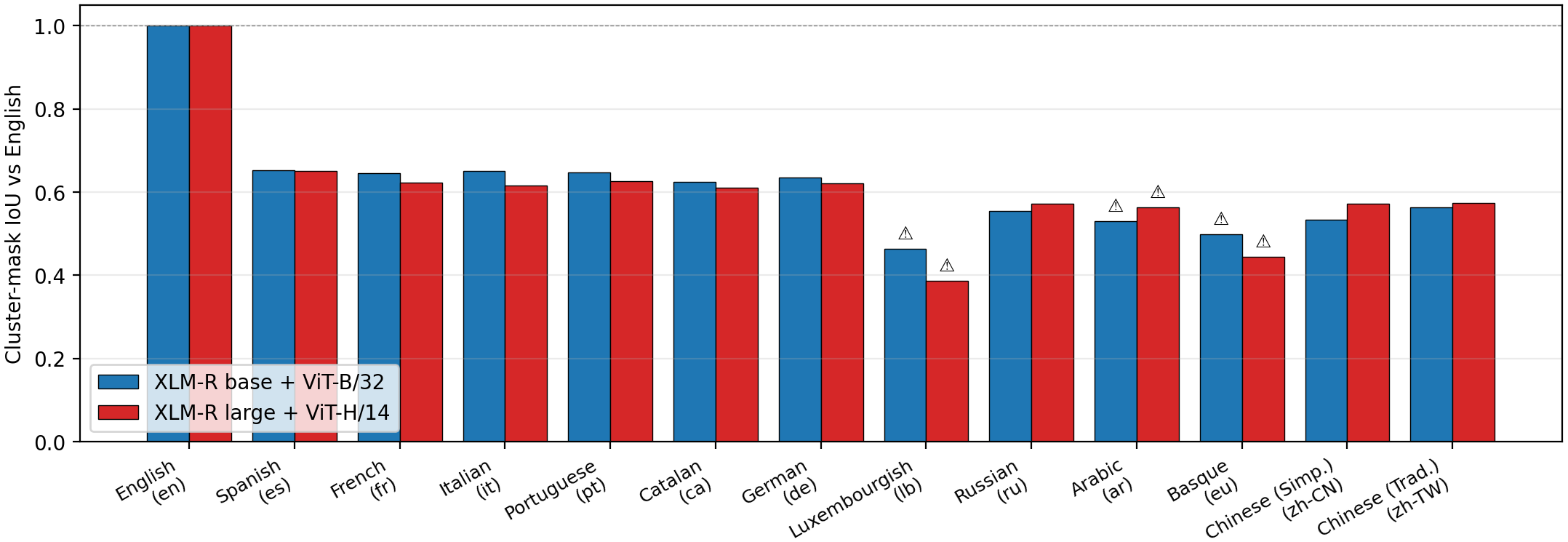}
  \caption{Per-language cluster-mask IoU against the English
    reference, for both CLIP backbones. The three low-resource
    languages ($\triangle$) consistently occupy the lower tail at both
    scales. HR\,$-$\,LR IoU gap = $+0.114$ (base), $+0.143$ (large);
    Wilcoxon HR\,$>$\,LR $p < 10^{-300}$ in both cases.}
  \label{fgr:figure1}
\end{figure}

Figure~\ref{fgr:figure1} reports per-language cluster-mask IoU against
the English reference for both backbones. The pattern is consistent
across architectures: every non-English language exhibits a substantial
drop in cross-language agreement, and the three low-resource languages
(Arabic, Basque, Luxembourgish, marked $\triangle$) consistently
occupy the lower tail of the distribution. On the smaller backbone the
HR\,$-$\,LR gap is $+0.1145$ in cluster-mask IoU
($0.6123$ vs $0.4978$); on the larger backbone the gap widens to
$+0.1428$ ($0.6076$ vs $0.4647$). Paired WILCOXON tests with the
one-sided alternative HR\,$>$\,LR yield $p < 10^{-300}$ at both backbone scales
on $n = 2{,}310$ paired (image, concept) observations per language,
together with significant FRIEDMAN tests across the twelve non-English
languages ($\chi^{2}_{\text{base}} = 8065.8$, $\chi^{2}_{\text{large}}
= 9478.4$, both $p < 10^{-300}$). 

On the three spatial-agreement metrics (cluster-mask IoU,
top-5\% IoU, Spearman~$\rho$), HR\,$>$\,LR is highly significant in
all twelve Wilcoxon and eighteen per-low-resource-language
MANN--WHITNEY comparisons (worst $p = 1.2 \times 10^{-86}$). On
peak similarity, by contrast, the HR/LR difference is small or
non-significant on the smaller backbone (Wilcoxon $p = 0.047$),
foreshadowing the mechanism diagnostic of Section~\ref{sn:o3}.

This pattern isolates the multilingual penalty to the text branch:
because the visual encoder is identical across languages by
construction (Section~\ref{sn:probe}), the cross-language disagreement
must originate downstream of the visual stage, in the
language-conditioned text encoding or in its alignment with the shared
visual embedding space. The persistence of the gap at both
architectural scales rules out an explanation that depends on a
specific visual backbone configuration.

\subsection{Scale Does Not Close the Low-Resource Gap}
\label{sn:o2}

A natural prediction is that increasing model capacity should narrow
the multilingual penalty, since larger encoders typically generalise
better across distributional shifts in the input. The data falsify
this prediction in the specific direction that matters most for
low-resource deployment.

\begin{figure}[t]
  \centering
  \includegraphics[width=\linewidth]{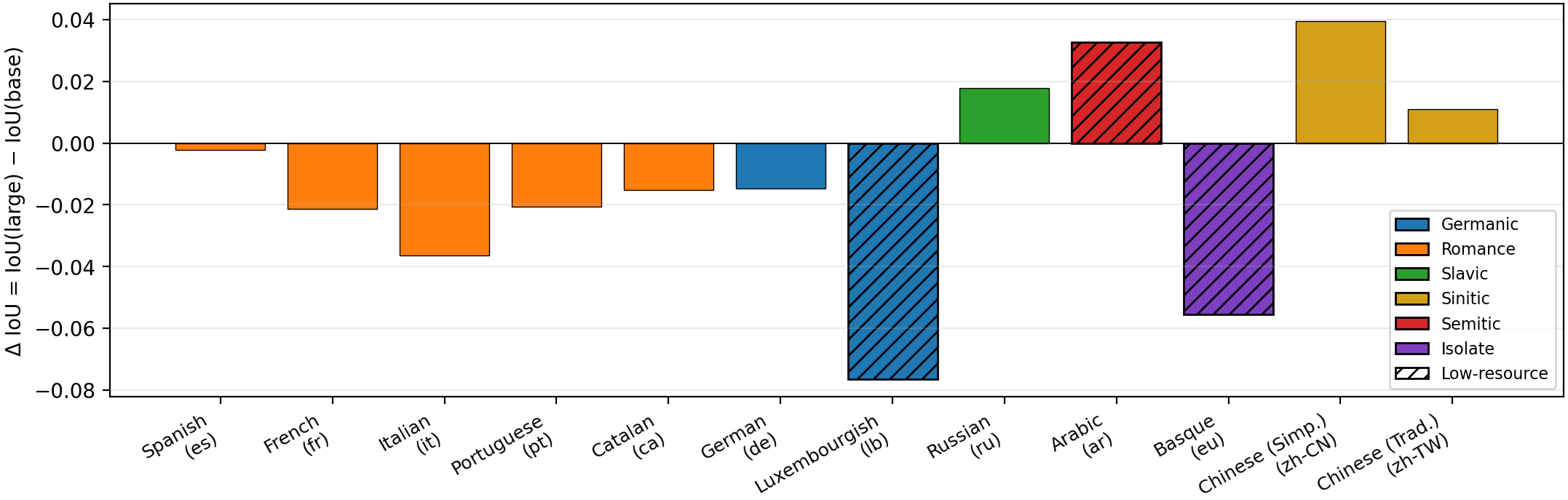}
  \caption{Per-language IoU shift under scaling,
    $\Delta_\text{IoU}^{(\ell)} =
    \mathrm{IoU}^{(\ell)}_\text{large} -
    \mathrm{IoU}^{(\ell)}_\text{base}$.
    Hatched bars mark low-resource languages. Basque
    ($\Delta = -0.056$) and Luxembourgish ($\Delta = -0.076$) lose
    cross-language agreement under scaling; Arabic ($\Delta = +0.033$)
    and Chinese (Simp., $\Delta = +0.039$) recover.}
  \label{fgr:figure3}
\end{figure}

Figure~\ref{fgr:figure3} reports the per-language IoU shift induced by
the $7\times$ visual-parameter scale increase, $\Delta_\text{IoU}^{(\ell)}
= \mathrm{IoU}^{(\ell)}_\text{large} - \mathrm{IoU}^{(\ell)}_\text{base}$.
Three regimes are visible. The high-resource Romance and Germanic
languages exhibit small negative shifts in the range
$-0.04 \leq \Delta \leq 0$ (Spanish $-0.002$, German $-0.015$,
Italian $-0.036$), consistent with a mild architectural mismatch
between the larger ViT-H/14 dense-feature recipe and the BDD100K
domain. Russian, Arabic, and the two Chinese variants exhibit modest
positive shifts in the range $+0.01 \leq \Delta \leq +0.04$
(Arabic $+0.033$, Chinese-Simplified $+0.039$). Basque and
Luxembourgish, by contrast, exhibit substantial \emph{negative}
shifts: $\Delta_{\text{eu}} = -0.056$ and $\Delta_{\text{lb}} = -0.076$
respectively, the two most negative shifts in the entire experiment.

The split among the three low-resource languages is informative. Arabic and Chinese, which are tokeniser-fertility-disadvantaged but
\emph{appear in substantial quantities} in LAION-5B's training corpus,
benefit from scale: a larger text encoder produces better cross-lingual
alignment for languages that are present in the pretraining
distribution. Basque and Luxembourgish, which are
under-represented at the corpus level rather than merely at the
tokeniser level, do not benefit from scale, and in fact lose ground.
This suggests that scaling the encoder cannot, on its own,
compensate for absent training data: it amplifies whatever signal is
present in the multilingual corpus, including the absence of signal
where coverage is thin. The HR\,$-$\,LR IoU gap therefore widens from
$+0.114$ to $+0.143$ between backbones rather than closing.

\subsection{The Mechanism Is Spatial Misalignment, Not Signal Collapse}
\label{sn:o3}

The peak-similarity ratio
$\eta^{(\ell)}(O, c) = \max S^{(\ell)}_c(O) / \max S^{(\ell_0)}_c(O)$
discriminates two failure modes of cross-language grounding.
\emph{Signal collapse} corresponds to $\eta \ll 1$: the
language-conditioned text encoder fails to produce strong activation
anywhere in the image. \emph{Spatial misalignment} corresponds to
$\eta \approx 1$ with low IoU: the encoder produces strong activation,
but in the wrong region of the image.

\begin{figure}[t]
  \centering
  \includegraphics[width=\linewidth]{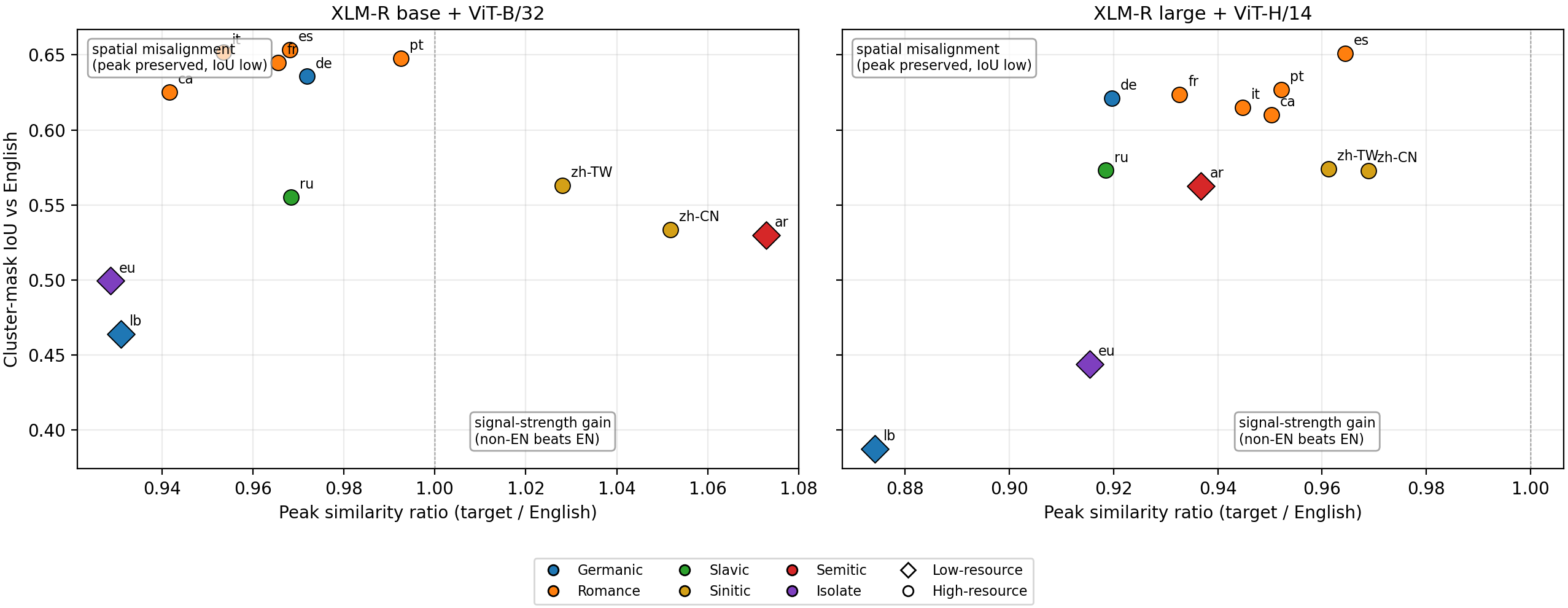}
  \caption{Mechanism diagnostic: peak-similarity ratio vs cluster-mask
    IoU. $\eta \approx 1$ with low IoU indicates spatial misalignment;
    $\eta \ll 1$ would indicate signal collapse.}
  \label{fgr:figure4}
\end{figure}

Figure~\ref{fgr:figure4} plots the per-language peak ratio against
cluster-mask IoU for both backbones, with low-resource languages
marked as filled diamonds. Peak ratios cluster near unity at both
scales (mean $0.981$ at base, $0.937$ at large), with no language
falling below $0.87$ even at the larger architecture. The drop in
cluster-mask IoU from English's value of $1.0$ down to
$0.39$--$0.65$ across non-English languages is therefore not driven by
a corresponding drop in peak signal: low-resource languages activate
on the wrong region rather than failing to activate at all.

The implication for downstream systems is concrete:
\textbf{confidence-based filtering on peak similarity is insufficient
to detect language-induced grounding errors.} A system that accepts
predictions when peak similarity exceeds a threshold will not flag
multilingual misalignment, because the peak is preserved.
Spatial-consistency or cross-language-agreement checks are needed
instead. We return to this design implication in
Section~\ref{sn:discussion}.

\subsection{Concept-Level Localisation of the Penalty}
\label{sn:o_concept}

\begin{figure}[t]
  \centering
  \includegraphics[width=\linewidth]{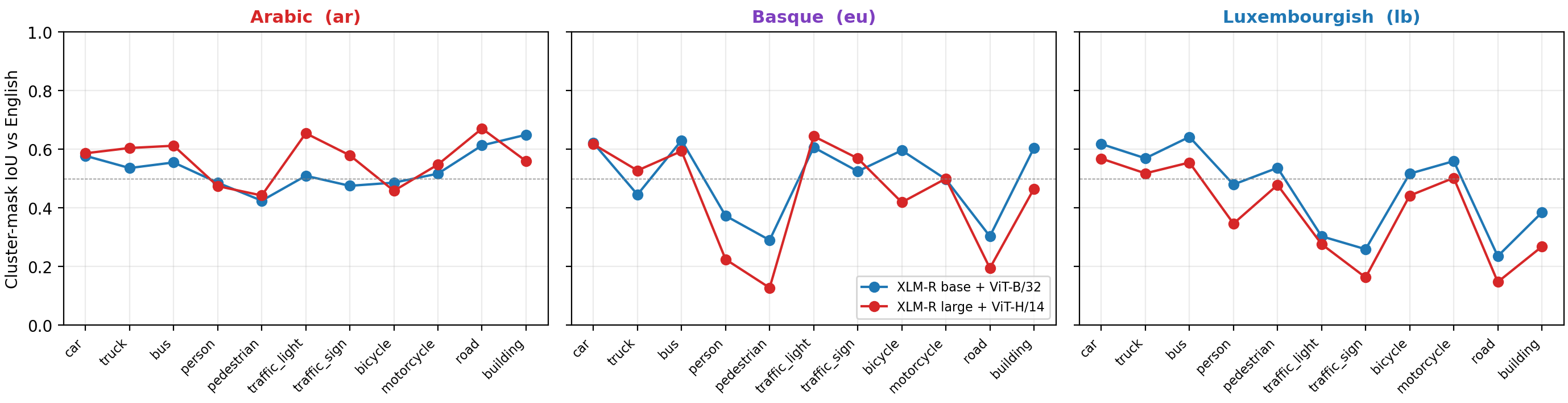}
  \caption{Per-concept cluster-mask IoU for the three low-resource
    languages under both backbones. Arabic uniformly improves under
    scaling; Basque and Luxembourgish exhibit concept-localised
    collapse, particularly on \emph{road}, \emph{pedestrian}, and
    \emph{traffic\_sign}.}
  \label{fgr:figure6}
\end{figure}

The aggregate statistics conceal substantial variation across
concepts. Figure~\ref{fgr:figure6} plots per-concept cluster-mask IoU
for the three low-resource languages under both backbones. The three
languages exhibit qualitatively different concept profiles. Arabic
recovers across nearly every concept under scaling, with the largest
gains on \emph{traffic\_light} ($0.51 \to 0.66$) and \emph{road}
($0.61 \to 0.67$). Luxembourgish degrades on every concept under
scaling, with the most severe drops on \emph{traffic\_sign}
($0.26 \to 0.16$) and \emph{road} ($0.24 \to 0.15$). Basque shows the
sharpest concept-localised collapse of any language in the study:
\emph{road} drops from $0.30$ to $0.19$ and \emph{pedestrian} from
$0.29$ to $0.13$ under scaling, while \emph{traffic\_light} and
\emph{building} are largely preserved.

The dominant pattern is that scaling \emph{redistributes} rather
than reduces low-resource error: gains on common, high-frequency
concepts (\emph{car}, \emph{building}) come at the cost of larger
losses on safety-critical concepts that are likely under-represented
in the multilingual pretraining corpus
(\emph{road}, \emph{pedestrian}, \emph{traffic\_sign} in Basque and
Luxembourgish). This concept-localised structure is not visible in
the language-aggregate IoU statistic of Section~\ref{sn:o1} and
motivates concept-aware reporting in future multilingual VLM
benchmarks.

\subsection{Energy}
\label{sn:o_energy}

Total inference for $30{,}030$ (image, concept, language) queries
consumed $116.15$\,Wh on the smaller backbone and $100.85$\,Wh on the
larger one, normalising to $E_{1\text{K}} = 3.87$ and $3.36$\,Wh per
$1{,}000$ queries respectively, at a mean GPU draw of $98.9$ and
$99.3$\,W on the NVIDIA H200 SXM. The slight reduction in $E_{1\text{K}}$ at the larger scale
reflects more efficient batched visual-encoder utilisation, which
amortises the per-image overhead more effectively than the smaller
ViT-B/32 configuration. Both figures are an order of magnitude below
typical autoregressive multilingual VLM inference budgets reported
under the same NVML protocol~\cite{aienergyscore2025,luccioni2024power}.
We discuss the resulting deployment implications in
Section~\ref{sn:discussion}.


\section{Discussion}
\label{sn:discussion}

We draw out four implications of the empirical findings: a
methodological observation about how multilingual visual grounding
should be \emph{measured}, an interpretation of the structure of the
penalty itself, two design implications for downstream multilingual
systems, and an energy--accuracy trade-off relevant to deployment.

\subsection{Symmetric Scoring as a Methodological Contribution}
\label{sn:disc_method}

Multilingual evaluation of foundation models has historically relied
on metrics that compare model outputs to English reference
material---keyword recall against English vocabularies, cosine
similarity to English embeddings, BLEU or ROUGE against
English-translated
ground-truth~\cite{ahuja2023mega,lai2023chatgpt,reimers2019sentence}.
Such choices conflate genuine cross-language perceptual disparity with
measurement bias toward English~\cite{wendler2024llamas}, and the
relative contribution of each cannot be recovered from the published
numbers. The dense-CLIP probe avoids this conflation by construction
(Section~\ref{sn:probe}): both maps live in the same dense embedding
space, are produced by the same visual encoder applied to the same
image, and are compared via metrics that are symmetric in their two
arguments. The penalty we measure is therefore attributable to the
multilingual encoder itself, not to the measurement framework. Any
future evaluation of multilingual dense grounding can adopt the same
symmetric protocol.

\subsection{Corpus Coverage versus Tokeniser Fertility}
\label{sn:disc_corpus_vs_tok}

The asymmetric response of the three low-resource languages to scaling
(Section~\ref{sn:o2}) separates two failure modes typically
discussed together. Tokeniser-fertility
disparity~\cite{rust2021good,petrov2024language} predicts
disadvantage at fixed compute that diminishes when the encoder gains
capacity to absorb longer token sequences. Arabic and Chinese are
consistent with this account: both improve under scaling
($\Delta_{\text{ar}} = +0.033$, $\Delta_{\text{zh-CN}} = +0.039$).
Basque and Luxembourgish show the opposite pattern---both \emph{lose}
agreement under scaling ($\Delta = -0.056, -0.076$), with
concept-localised collapse on safety-critical nouns
(Section~\ref{sn:o_concept}). This is consistent not with fertility
but with corpus-coverage failure: when a language is largely absent
from the multilingual training
distribution~\cite{schuhmann2022laion5b}, additional encoder capacity
has no signal to amplify, and may amplify the misalignment that the
smaller encoder partially smoothed. Scale alone cannot rescue
languages absent from the pretraining corpus; targeted multilingual
data curation is required.

\subsection{Implications for Downstream Systems}
\label{sn:disc_design}

The spatial-misalignment finding (Section~\ref{sn:o3}) has a direct
consequence for system design: confidence-based filtering on peak
similarity~\cite{zhou2022maskclip,busch2025onemap} cannot detect
language-induced grounding errors, because peak similarity is
preserved at near-English levels (mean ratio $0.94$ at the larger
scale) while the location of activation shifts. A complementary filter
follows from the same observation: cross-language agreement itself
can serve as a runtime confidence signal. A system that issues the same query in two or more languages, then rejects predictions where the resulting maps disagree spatially, detects misalignments that single-language peak filtering cannot. This pattern fits naturally
into cooperative-perception
architectures~\cite{LLMfusiondeCurto2025} where multiple
language-conditioned queries are fused at runtime to produce a
consensus grounding.

\subsection{Energy and Limitations}
\label{sn:disc_energy_limits}

\begin{figure}[t]
  \centering
  \includegraphics[width=\linewidth]{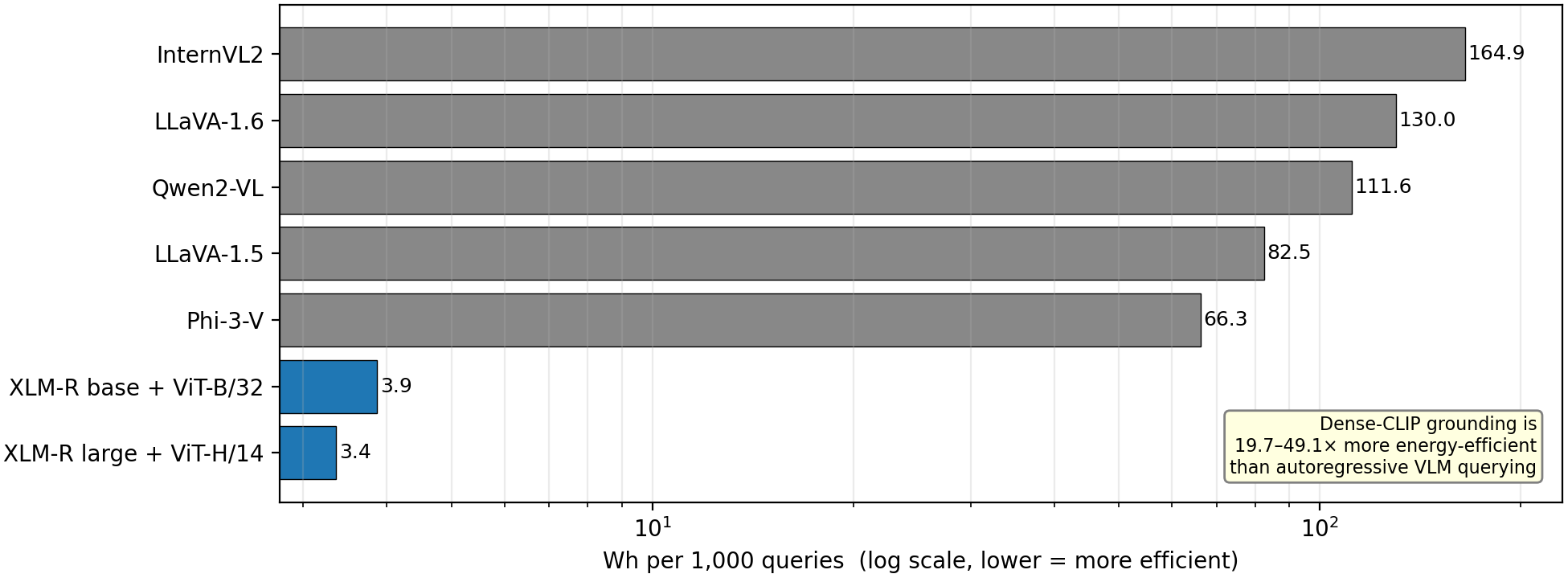}
  \caption{Inference energy per $1{,}000$ queries on a log scale,
    measured under a uniform NVML 10\,Hz protocol. Dense-CLIP grounding
    operates roughly $20$--$50\times$ below the per-query budget of
    autoregressive multilingual VLMs reported under the same protocol
    in~\cite{aienergyscore2025,luccioni2024power}.}
  \label{fgr:figure7}
\end{figure}

Figure~\ref{fgr:figure7} places the probe on a uniform energy scale
alongside autoregressive multilingual VLM budgets reported under the
same NVML protocol~\cite{aienergyscore2025,luccioni2024power}. At
$3.4$--$3.9$\,Wh per $1{,}000$ queries, dense-CLIP grounding sits
roughly $20$--$50\times$ below typical generative VLM inference, with
margin sufficient to absorb the cross-language consistency check
proposed above. The substrate is not a substitute for generative VLMs in tasks that require free-form reasoning, but for the dense-grounding subtask it is markedly cheaper and more directly auditable across languages. 

Three caveats bound the scope of these claims. The probe operates
on a fixed $210$-image BDD100K subset~\cite{yu2020bdd100k}, so
concept-level findings are conditional on this distribution.
Single-noun queries do not exercise compositional or relational
grounding (\emph{the car next to the bus}), where the penalty may
differ in magnitude or mechanism. Energy figures are recorded on a
single server GPU; the absolute Wh per $1{,}000$ queries will shift on
edge accelerators, though the relative ranking between contrastive and
autoregressive grounding is hardware-robust.


\section{Conclusion}
\label{sn:conclusion}

We presented a controlled multilingual probe of dense CLIP grounding
that holds the visual encoder identical across thirteen languages
and varies only the XLM-RoBERTa text branch. Across two architectural
scales spanning a $7\times$ visual-parameter gap and $n = 2{,}310$
paired observations per non-English language, the multilingual penalty
persists at both scales (HR\,$-$\,LR cluster-mask IoU gap $+0.114$ and
$+0.143$; Wilcoxon $p < 10^{-300}$), localising it to the text branch.
Scaling the encoder \emph{widens} the gap for languages
under-represented in the pretraining corpus (Basque, Luxembourgish)
while improving tokeniser-disadvantaged ones (Arabic, Chinese),
separating two failure modes the literature has often conflated. Peak
similarity is preserved across languages while the spatial location of
activation shifts, ruling out signal collapse and identifying spatial
misalignment as the dominant failure mode---with the immediate
consequence that confidence-based filtering on peak similarity cannot
detect language-induced grounding errors. The probe operates at
$3.4$--$3.9$\,Wh per $1{,}000$ queries, roughly $20$--$50\times$ below
typical autoregressive multilingual VLM inference, making
cross-language agreement a viable runtime confidence signal in
energy-constrained settings. Scale and tokeniser improvements are not
interchangeable remedies: languages absent from the pretraining
corpus require targeted data curation, not larger encoders. Extensions to compositional queries, additional languages, and edge-accelerator deployment scenarios are natural directions for follow-up work.

\section*{Acknowledgements}
This research was supported by the LUXEMBOURG Institute of
Science and Technology through the projects
``ADIALab-MAST'' and ``LLMs4EU''
(Grant Agreement No~101198470) and the BARCELONA
Supercomputing Center through the project ``TIFON''
(File number MIG-20232039). Mauro Liz would also like to thank Universidad Pontificia Comillas for the opportunity to participate in the international exchange program with the Department of Electrical and Computer Engineering at Boston University.

\section*{Code Availability}
The implementation required to reproduce the experimental
analysis and compute the AI Energy Score metrics presented
in this paper is publicly available at:
\url{https://github.com/drdecurto/clip_multilingual_analysis}

\bibliographystyle{IEEEtran}
\bibliography{sample.bib}

\end{document}